\documentclass{article} 
\usepackage{iclr2025_conference,times}

\usepackage{amsmath,amsfonts,bm}

\def\eqref#1{equation~\ref{#1}}

\def\1{\bm{1}}

\DeclareMathAlphabet{\mathsfit}{\encodingdefault}{\sfdefault}{m}{sl}
\SetMathAlphabet{\mathsfit}{bold}{\encodingdefault}{\sfdefault}{bx}{n}

\usepackage{hyperref}
\usepackage{url}
\usepackage{graphicx}

\usepackage{booktabs} 
\usepackage{array}    
\usepackage{graphicx} 
\usepackage{siunitx}  
\usepackage{caption}  

\title{XAMBA: Enabling Efficient State Space \\ Models on Resource-Constrained Neural Processing Units}

\author{Arghadip Das$^{1}$\thanks{Corresponding author.  \texttt{das169@purdue.edu}}~, Arnab Raha$^{2}$, Shamik Kundu$^{2}$, Soumendu Kumar Ghosh$^{2}$,\\ \textbf{Deepak Mathaikutty}$^{2}$,
    \textbf{and Vijay Raghunathan}$^{1}$ \\
    $^{1}$Electrical and Computer Engineering, Purdue University \\
    $^{2}$Advanced Architecture Research Team, NPU IP, CGAI (CCG), Intel Corporation\\
    \texttt{\{das169, vr\}@purdue.edu} \\
    \texttt{\{arnab.raha, shamik.kundu\}@intel.com} \\
    \texttt{\{soumendu.ghosh, deepak.a.mathaikutty\}@intel.com} 
}

\iclrfinalcopy

\begin{document}

\maketitle

\begin{abstract}
State-Space Models (SSMs) have emerged as efficient alternatives to transformers for sequential data tasks, offering linear or near-linear scalability with sequence length, unlike transformers with quadratic-complexity attention. This makes SSMs ideal for long-sequence tasks in natural language processing (NLP), vision, and edge AI applications such as real-time transcription, translation, and contextual search. These applications demand lightweight, high-performance models for deployment on resource-constrained devices like laptops and PCs. 
\textcolor{black}{While specialized accelerators have been proposed for emerging neural networks, designing new hardware is time-intensive, costly, and impractical for every model. Instead, optimizing models for existing neural processing units (NPUs) in AI PCs offers a scalable and efficient solution.
Towards this end, we propose \textbf{XAMBA}, \textit{the first framework to enable and optimize SSMs on commercial off-the-shelf (COTS) state-of-the-art (SOTA) NPUs}.}
Our approach follows a systematic three-step methodology: (1) enabling SSMs on NPUs, (2) optimizing performance to meet target Key Performance Indicator (KPI) requirements, and (3) trading accuracy for additional performance gains. After enabling SSMs on NPUs, XAMBA addresses key performance bottlenecks with two techniques: \textbf{CumBA} and \textbf{ReduBA}. These replace sequential CumSum and ReduceSum operations with matrix-based computations, significantly improving execution speed and memory efficiency. In addition, \textbf{ActiBA} further enhances performance by mapping computationally expensive activation functions (\emph{e.g.}, Swish, Softplus) to NPU hardware using piecewise linear approximations, reducing latency with minimal accuracy loss. Experimental evaluations on an Intel\textregistered\ Core\texttrademark\ Ultra Series 2 AI PC demonstrate that XAMBA achieves significant performance improvements, reducing execution latency by up to 4.8$\times$ compared to baseline implementation. Our code implementation is available at \href{https://github.com/arghadippurdue/XAMBA}{this link}.

\end{abstract}

\section{Introduction}\label{sec_intro}
\begin{figure}[t!]
\begin{center}
\includegraphics[width=\columnwidth]{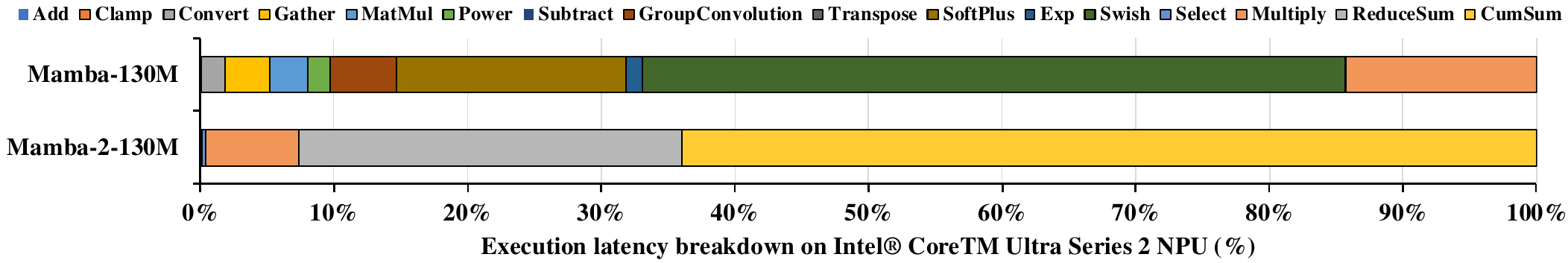}
\end{center}
\caption{Execution bottlenecks for Mamba and Mamba-2 on Intel\textregistered\ Core\texttrademark\ Ultra Series 2 NPU. Mamba is limited by sequential DSP execution of Swish (a.k.a. SiLU) and SoftPlus, while Mamba-2 faces CumSum and ReduceSum bottlenecks. Refer ~\ref{subsec:mamba1_vs_2} for models' architectural details.}
\label{fig:motivation_exec_lat_brkdwn}
\end{figure}

SSMs, a classical framework for modeling dynamic systems through first-order differential equations, have gained prominence in machine learning for their efficiency in handling sequential data. Unlike transformers, which rely on quadratic-complexity attention mechanisms, SSMs achieve linear or near-linear scalability with sequence length by leveraging principles from convolutional and recurrent neural networks~\cite{s4}. This makes SSMs highly suitable for long-sequence tasks~\cite{mambaextend} like natural language processing, computer vision, and medicine, where they match transformers' modeling capabilities with significantly lower computational overhead. Their efficiency is particularly critical for edge applications, such as personal assistants and real-time transcription, where SSMs enable transformative AI with reduced resource consumption and improved energy efficiency.
Among SSM-based architectures, Mamba~\cite{mamba} and Mamba-2~\cite{mamba2} are the most prominent.
Mamba introduces algorithmic innovations, such as selective scan, which efficiently processes sequential data by selectively updating only relevant states. Mamba-2 further refines this approach with the structured state-space duality (SSD) framework, which connects SSMs with attention mechanisms, enabling the reuse of transformer optimization techniques~\cite{mamba360}. These algorithmic advancements position SSMs as strong candidates for replacing transformers in resource-constrained environments.
\textcolor{black}{\textit{While specialized accelerators exist for emerging networks, designing a new hardware is costly and impractical for every model. Instead, \textbf{XAMBA} repurposes NPUs for SSMs, enabling efficient deployment on existing hardware.}}
The efficiency of SSMs makes them ideal for deployment on NPUs, specialized accelerators optimized for data-parallel operations like matrix multiplication. Modern edge processors, such as AI PCs from Intel, Qualcomm, and AMD, integrate NPUs alongside CPUs and GPUs to support diverse AI workloads. NPUs, designed for high throughput and energy efficiency, consist of two key components: (1) \textit{Multiply-and-accumulate (MAC) Processing Units (MPUs)}~\cite{flexnn}, which handle parallelized matrix operations, and (2) \textit{Digital Signal Processors (DSPs)}, which execute sequential tasks like non-linear activations and cumulative summations. This architecture makes NPUs ideal for continuous, resource-intensive workloads such as personal assistants, real-time transcription, and contextual search.
Prior work has demonstrated the benefits of deploying transformer-based models (\textit{e.g.}, Llama, Phi) on NPUs, achieving significant improvements in inference latency and energy efficiency~\cite{llm_npu}. However, transformers' quadratic complexity limits their scalability for long sequences. SSMs, with their linear complexity, offer a natural fit for NPUs, enabling even greater performance and energy efficiency. Despite this, deploying SSMs on NPUs presents unique challenges. Their sequential computations (\textit{e.g}., CumSum, Swish, Softplus) and specialized operators misalign with NPUs' data-parallel architecture, leading to inefficient DSP execution, increased memory traffic, and underutilized parallel units. While prior work has optimized recurrent models like LSTMs on NPUs~\cite{rnn_npu} and proposed specialized accelerators like MARCA for SSMs~\cite{marca}, \textit{no prior work addresses the challenges of deploying SSMs on COTS SOTA NPUs.}
As shown in Figure~\ref{fig:motivation_exec_lat_brkdwn}, SSMs face significant bottlenecks on NPUs. For Mamba, activation functions like Swish and Softplus~\cite{openvino_ops} dominate execution time due to inefficient DSP execution. In Mamba-2, CumSum and ReduceSum emerge as critical bottlenecks, exacerbated by poor memory reuse and increased off-chip traffic. These challenges underscore the need for specialized optimizations to fully leverage SSMs on NPUs.
To address these challenges, we propose \textbf{XAMBA}, \textit{the first work to enable and optimize SSMs on COTS NPUs.} XAMBA follows a systematic 3-step methodology: (1) enabling SSMs on NPUs, (2) optimizing performance to meet target KPI requirements, and (3) trading accuracy for additional performance gains. 


\begin{itemize}
    \item \textbf{Optimizing SSM Performance on NPUs}: After enabling SSMs on NPUs, XAMBA introduces \textbf{CumBA} and \textbf{ReduBA} to address key bottlenecks in SSM execution on NPUs. CumBA replaces sequential CumSum operations with matrix multiplication (MatMul) using a precomputed mask, leveraging the high-frequency MPUs for faster execution. ReduBA computes ReduceSum via matrix-vector multiplication (MVM) with a precomputed vector mask, further reducing latency. Both techniques enhance memory efficiency by increasing data reuse and reducing SRAM access, while CumBA leverages sparsity in its mask for additional optimizations using Zero Value Compression (ZVC) and compute skip.

    \item \textbf{Trading Accuracy for Performance Gains}: XAMBA introduces \textbf{ActiBA}, which maps computationally expensive activation functions (e.g., Swish, Softplus) onto the NPU's Piecewise Linear Unit (PLU) using Configurable Lookup Table (C-LUT) during the drain phase of the previous layer. ActiBA uses piecewise linear approximations to implement these functions with negligible accuracy loss, avoiding sequential DSP execution and reducing memory access overhead.

    \item XAMBA achieves significant latency reductions for Mamba and Mamba-2 on an Intel\textregistered\ Core\texttrademark\ Ultra Series 2 NPU. CumBA reduces execution latency by 2.7$\times$, ReduBA by 1.2$\times$, and ActiBA by up to 2.6$\times$ compared to baseline unoptimized implementation.
\end{itemize}

\section{XAMBA Design Methodology}\label{sec_xamba_design}

\begin{figure}[t!]
\begin{center}
\includegraphics[width=\columnwidth]{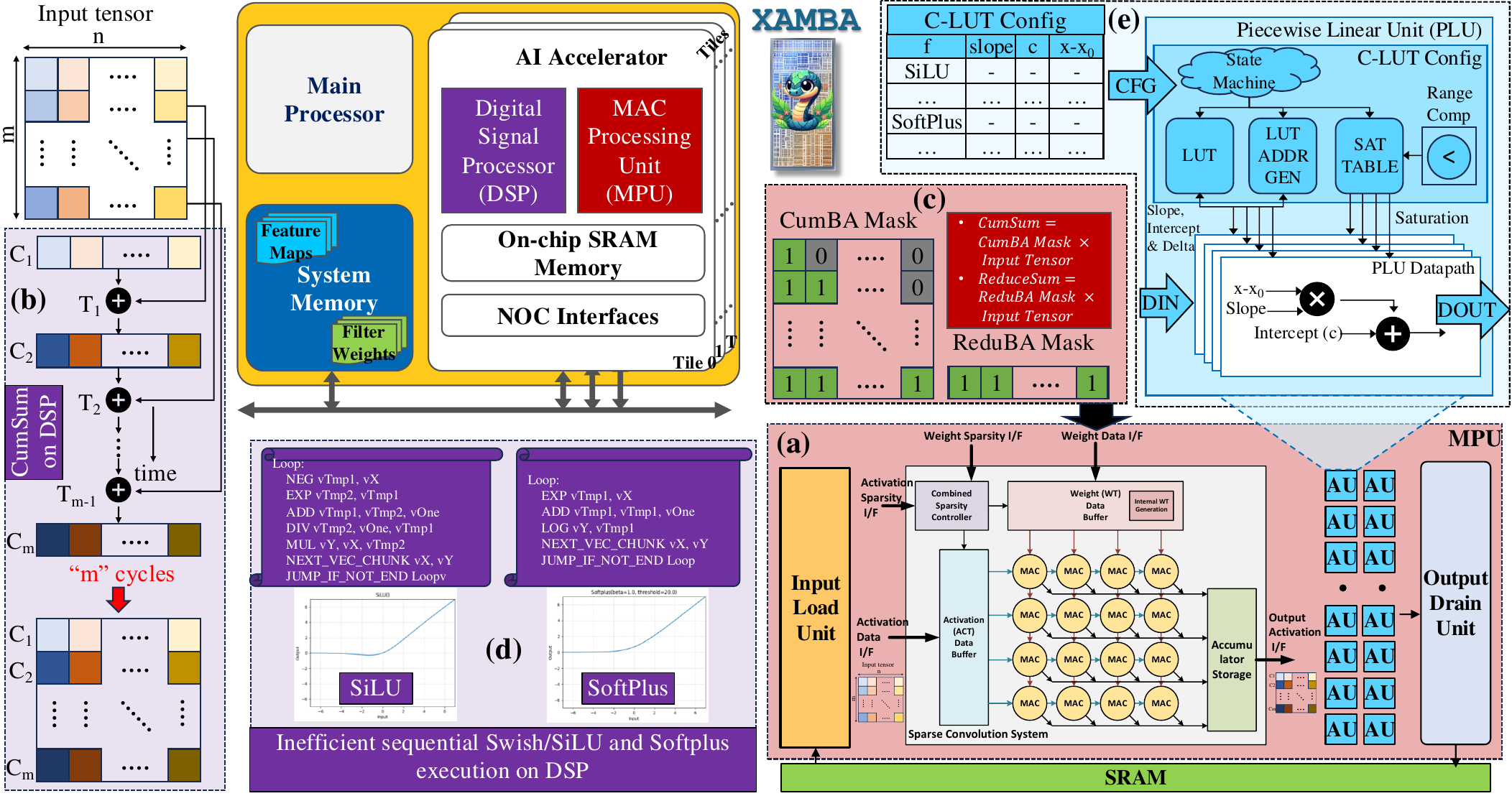}
\end{center}
\caption{XAMBA:
(a) NPU architecture
(b) Sequential CumSum and ReduceSum computation on a DSP.
(c) CumBA and ReduBA masks for optimized computations.
(d) Sequential execution of activation functions (Swish/SiLU and SoftPlus) on DSP.
(e) ActiBA: Efficient execution of SoftPlus and Swish activations using C-LUT in PLU.}\label{fig:xamba}
\end{figure}

Before detailing the design methodology, it is essential first to understand the underlying system architecture (refer Figure~\ref{fig:xamba}). We consider an output-stationary MPU architecture (as shown in Figure~\ref{fig:xamba}(a)) inspired by ~\cite{flexnn}. 
Although our case study considers an output-stationary MPU architecture, the proposed techniques are generic and can be applied to other NPUs without loss of generality. \textbf{Step-1:} The \textit{first step} of XAMBA is to enable SSMs on NPUs. Since NPUs generally support static input shapes, we use a prefill model with a fixed number of input tokens to generate the hidden states, applying padding for smaller inputs. For subsequent token generation, we employ a separate model that uses the cached hidden states to ensure efficiency. 

\subsection{Step-2: Optimizing SSM Performance to meet target KPI requirements}

\textbf{CumBA:}  
As highlighted in Figure~\ref{fig:motivation_exec_lat_brkdwn}, one of the major bottlenecks in executing Mamba-2 on NPUs is the CumSum operation.
\textcolor{black}{A deeper analysis reveals that Mamba-2 contains three CumSum operations per block (Figure ~\ref{fig:mamba_vs_mamba2}), but the primary bottleneck, denoted as $CumSum_b$, accounts for more than 99.9\% of the total CumSum execution time. This bottleneck arises in step-1 of the SSD framework of Mamba-2 (Listing 1 of ~\cite{mamba2}).  
Within SSD, the input sequence is split into blocks, and step 1 computes the intra-chunk output, assuming an initial state of zero. $CumSum_b$, appearing at the start of this step, computes semi-separable (SS) matrices essential for modeling long-range dependencies across input segments.
The bottleneck stems from the large matrix dimensions associated with $CumSum_b$: In Mamba-2 130M, $CumSum_b$ operates on a $256 \times 256$ matrix, whereas the other CumSum operations in the model involve significantly smaller dimensions ($256$ and $2 \times 2$). }
Figure~\ref{fig:xamba}(b) depicts that executing CumSum on NPUs leads to high latency due to its \textit{sequential nature} on the DSP. Given an input tensor $\mathbf{X} \in \mathbb{R}^{m \times n}$, the standard CumSum operation along the row dimension is expressed as $\mathbf{C}_{i,j} = \sum_{k=1}^{i} \mathbf{X}_{k,j}$ for all $i \in [1, m]$, $j \in [1, n]$. This requires $m$ sequential cycles, assuming the DSP has an $n$-width vector adder. CumSum is processed in smaller chunks for higher-dimensional tensors exceeding register file capacity, requiring frequent on-chip SRAM memory transfers. This increases latency, memory traffic, and bandwidth consumption, leading to inefficient data reuse and performance degradation.

To address these inefficiencies, XAMBA introduces \textbf{CumBA}, which transforms CumSum into a MatMul, leveraging the parallel processing capabilities of the NPU’s MPU. Specifically, CumBA precomputes (at compile-time) a lower triangular mask $\mathbf{M}_{\text{CumBA}}$ (Figure~\ref{fig:xamba}(c)), where $\mathbf{M}_{\text{CumBA}}(i, j) = 1$ if $j \leq i$ and $0$ otherwise. This enables CumSum to be computed as $\mathbf{C} = \mathbf{M}_{\text{CumBA}} \cdot \mathbf{X}$. By remapping CumSum to matrix multiplication, CumBA enables parallel execution, leveraging the MPU’s high-frequency MAC array to perform multiple operations simultaneously. It improves data reuse by utilizing the MPU’s larger local register files, reducing redundant memory reads/writes to SRAM compared to DSP-based execution. The MPU processes MatMul in a tiled manner, further improving data reuse and minimizing costly on-and-off-chip memory transfers. 

\begin{figure}[t!]
\begin{center}
\includegraphics[width=\columnwidth]{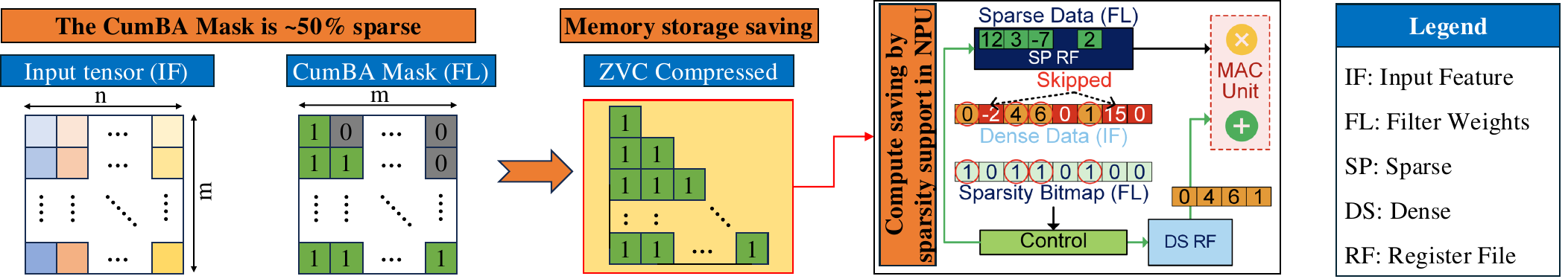}
\end{center}
\caption{CumBA: Enhancing memory, bandwidth, and compute efficiency by exploiting CumBA mask sparsity and two-sided sparsity acceleration in the NPU datapath.}\label{fig:cumba_zvc}
\end{figure}

\textit{CumBA memory savings using Zero Value Compression (ZVC):}
XAMBA applies ZVC~\cite{zvc} to compress the CumBA mask, a lower triangular binary matrix with $\sim$50\% zeros, significantly reducing storage and memory traffic (Figure~\ref{fig:cumba_zvc}). This compression leads to substantial memory savings, as only non-zero elements are stored. Additionally, modern NPUs utilize sparsity bitmaps to skip zero-value computations, further improving execution speed and energy efficiency. While CumBA benefits from ZVC-driven optimizations, the weights in Mamba and Mamba-2 exhibit minimal inherent sparsity, limiting acceleration gains from sparsity-aware execution. As future work, we plan to explore structured sparsity techniques in SSMs to further enhance NPU efficiency.

\textbf{ReduBA:} 
As illustrated in Figure~\ref{fig:motivation_exec_lat_brkdwn}, another significant bottleneck in the execution of Mamba-2 on NPUs is the ReduceSum operation.
\textcolor{black}{A detailed analysis shows that these bottlenecks originate from the reduction sum operations present in every step of the SSD framework in Mamba-2 (Listing 1 of ~\cite{mamba2}).}
Similar to CumSum, the ReduceSum operation suffers from high latency due to sequential DSP execution, as illustrated in Figure~\ref{fig:xamba}(b). Given an input matrix $\mathbf{X} \in \mathbb{R}^{m \times n}$, the ReduceSum along the row dimension is defined as $\mathbf{R}_{j} = \sum_{i=1}^{m} \mathbf{X}_{i,j} = \mathbf{C}_{m,j}$ for all $j \in [1, n]$.  

To mitigate this, XAMBA introduces \textbf{ReduBA}, which reformulates ReduceSum as a matrix-vector multiplication (MVM) using a precomputed vector mask $\mathbf{M}_{\text{ReduBA}}$ (Figure~\ref{fig:xamba}(c)), where $\mathbf{M}_{\text{ReduBA}}(i) = 1$ for all $i$. The ReduceSum operation is then computed as $\mathbf{R} = \mathbf{M}_{\text{ReduBA}} \cdot \mathbf{X}$. 
ReduBA improves upon CumBA by reusing the ReduBA vector mask $\mathbf{M}_{\text{ReduBA}}$ across all operations, reducing memory traffic. Unlike CumBA's matrix-matrix operations, where each computation fetches a new part of the mask, ReduBA’s matrix-vector multiplication applies the same mask repeatedly, minimizing memory accesses and optimizing bandwidth. ReduBA also leverages multiple MAC units in the MPU and a tiled computation strategy, further enhancing data reuse and reducing on-chip memory accesses, resulting in lower latency and improved memory efficiency for ReduceSum operations in Mamba-2 on NPUs.

\subsection{Step-3: Trading Accuracy for Additional Performance Gains}

\textbf{ActiBA:}  
As illustrated in Figure~\ref{fig:motivation_exec_lat_brkdwn}, two of the most significant bottlenecks in Mamba’s execution on NPUs are the Swish (a.k.a. SiLU) and Softplus activation functions. They introduce significant execution overhead when processed sequentially on the DSP, as depicted in Figure~\ref{fig:xamba}(d). 
The SiLU function is defined as $\text{SiLU}(x) = x \cdot \sigma(x)$ with $\sigma(x) = \frac{1}{1 + e^{-x}}$, while Softplus is given by $\text{Softplus}(x) = \frac{1}{\beta} \log (1 + e^{\beta x})$. Both functions exhibit nonlinearity near the origin but transition to linear behavior elsewhere, making them suitable for approximation using piecewise linear functions.

XAMBA introduces \textbf{ActiBA}, which leverages the Piecewise Linear Unit (PLU) found in modern NPUs to efficiently map the Swish and Softplus activation functions. The PLU, located within the Arithmetic Unit (AU) in the MPU's drain path, integrates a C-LUT, as shown in Figure~\ref{fig:xamba}(e).
The C-LUT stores precomputed slopes and intercepts for linear segments, enabling the approximation $f(x) \approx m_k x + c_k$ over intervals $[x_k, x_{k+1}]$. During runtime, the activation function is evaluated directly using the C-LUT, avoiding sequential DSP execution and significantly reducing latency. ActiBA also utilizes \textit{vertical fusion} by performing activation computations during the drain phase, which eliminates the need for storing and reloading intermediate outputs. This reduces memory access overhead and optimizes memory bandwidth usage. The simple linear computations, integrated into the data drain process, further minimize execution latency. By addressing both computational and memory inefficiencies, ActiBA drastically lowers end-to-end latency for Mamba-based models with \textit{negligible loss in quality}~\cite{dse_act}. Increasing the number of linear segments in the non-linear section of the activation functions can further reduce this loss without significantly impacting performance~\cite{flex_sfu}.

\section{Experimental Methodology}\label{sec_expt_meth}
\textbf{Networks and Datasets:}  
Experiments use pretrained state-space models from HuggingFace, specifically \texttt{mamba-130m-hf} and \texttt{mamba2-130m-hf}, with fixed input tokens of 4.
\textbf{Preprocessing and Conversion:}  
The models are converted from PyTorch to ONNX. They are then converted to OpenVINO~\cite{openvino} IR files (compressing weights to FP16 precision) and compiled into a binary using the OpenVINO NPU compiler. CumBA and ReduBA optimizations are applied during conversion, and ActiBA is emulated by replacing activation functions with ReLU.
\textbf{Platform:}  
Experiments run on an Intel\textregistered\ Core\texttrademark\ Ultra Series 2~\cite{lnl} platform (ASUS Zenbook S 14, 16GB RAM, 256V NPU).
\textbf{Performance Evaluation:}  
Models are evaluated using OpenVINO’s \texttt{benchmark\_app} tool, focusing on inference latency, with optimizations (CumBA, ReduBA, ActiBA) affecting NPU execution efficiency.
\textit{All results were collected using public frameworks (OpenVINO, PyTorch) and are replicable via the provided code (see abstract).}

\section{Results}\label{sec_results}
\begin{figure}[t!]
\begin{center}
\includegraphics[width=\columnwidth]{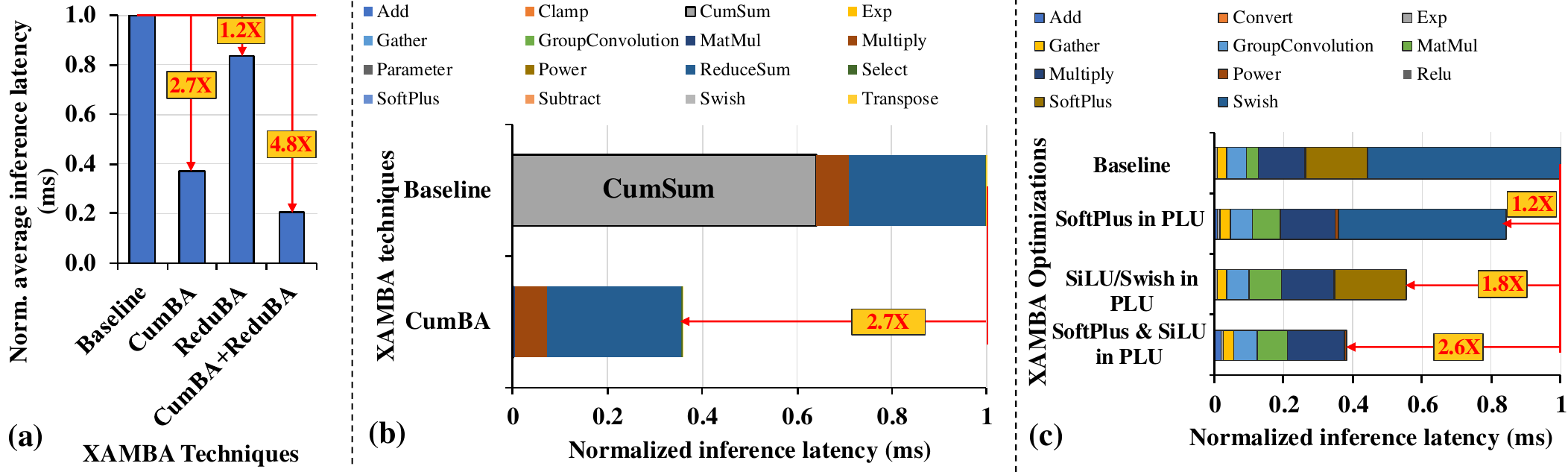}
\end{center}
\caption{Latency reduction for Mamba and Mamba-2 130M models on Intel\textregistered\ Core\texttrademark\ Ultra Series 2~\cite{lnl} NPU with XAMBA optimizations (CumBA, ReduBA, and ActiBA).}\label{plot:result}
\end{figure}

\begin{table}[t]
\caption{\small \textcolor{black}{Quality of XAMBA model variants across benchmarks. Arrows indicate direction of better performance. Groups are separated by model size.}}
\label{tab:mamba_results}
\centering
\scriptsize
\setlength{\tabcolsep}{3.5pt}
\begin{tabular}{@{}l S[table-format=1.2] *{6}{S[table-format=2.1]} S[table-format=2.1]@{}}
\toprule
\textbf{Model} & \textbf{\shortstack{LAMBADA\\PPL $\downarrow$}} & \textbf{\shortstack{HellaSwag\\ACC $\uparrow$}} & \textbf{\shortstack{PIQA\\ACC $\uparrow$}} & \textbf{\shortstack{ARC-E\\ACC $\uparrow$}} & \textbf{\shortstack{ARC-C\\ACC $\uparrow$}} & \textbf{\shortstack{OpenbookQA\\ACC $\uparrow$}} & \textbf{\shortstack{Winogrande\\ACC $\uparrow$}} & \textbf{\shortstack{AVG.\\ACC $\uparrow$}} \\
\midrule
Mamba-130M & 16.07 & 24.23 & 35.25 & 51.85 & 47.98 & 28.40 & 64.47 & 42.03 \\
Mamba-130M-PLU SiLU \& Softplus & {---} & 25.34 & 31.83 & 52.33 & 47.14 & 28.20 & 59.19 & 40.67 \\
Mamba2-130M & 16.79 & 24.15 & 35.27 & 52.80 & 47.39 & 30.60 & 64.91 & 42.52 \\
Mamba2-130M-PLU SiLU \& SoftPlus & 16.78 & 24.15 & 35.27 & 52.80 & 47.43 & 30.60 & 64.91 & 42.53 \\
\midrule
Mamba-370M & 8.14 & 27.90 & 46.48 & 52.09 & 54.97 & 30.80 & 69.48 & 46.95 \\
Mamba-370M-PLU SiLU \& Softplus & 8.13 & 27.99 & 46.49 & 52.01 & 54.92 & 30.80 & 69.48 & 46.95 \\
Mamba2-370M & 7.98 & 26.71 & 46.94 & 53.67 & 54.84 & 32.40 & 70.51 & 47.51 \\
Mamba2-370M-PLU SiLU \& SoftPlus & 7.98 & 26.71 & 46.94 & 53.83 & 54.84 & 32.40 & 70.51 & 47.54 \\
\midrule
Mamba-790M & 6.01 & 29.35 & 55.07 & 56.51 & 61.24 & 34.20 & 72.14 & 51.42 \\
Mamba-790M-PLU SiLU \& Softplus & 6.02 & 29.44 & 55.06 & 56.35 & 61.24 & 34.00 & 72.14 & 51.37 \\
Mamba2-780M & 5.85 & 28.50 & 54.91 & 56.91 & 60.98 & 36.20 & 72.03 & 51.59 \\
Mamba2-780M-PLU SiLU \& SoftPlus & 5.85 & 28.50 & 54.92 & 56.91 & 61.03 & 36.20 & 72.03 & 51.60 \\
\midrule
Mamba-1.4B & 5.04 & 32.94 & 59.11 & 57.30 & 65.53 & 36.40 & 74.16 & 54.24 \\
Mamba-1.4B-PLU SiLU \& Softplus & 5.04 & 32.59 & 59.07 & 57.22 & 65.45 & 36.40 & 74.16 & 54.15 \\
Mmaba2-1.3B & 5.02 & 33.11 & 59.94 & 58.25 & 64.14 & 37.80 & 73.23 & 54.41 \\
Mmaba2-1.3B-PLU SiLU \& SoftPlus & 5.02 & 33.19 & 59.93 & 58.33 & 64.14 & 37.80 & 73.23 & 54.44 \\
\midrule
Mamba-2.8B & 4.23 & 36.26 & 66.16 & 59.98 & 69.65 & 39.60 & 75.24 & 57.82 \\
Mamba-2.8B-PLU SiLU \& Softplus & 4.23 & 36.26 & 66.16 & 60.14 & 69.61 & 39.60 & 75.24 & 57.84 \\
Mmaba2-2.7B & 4.09 & 36.26 & 66.58 & 61.72 & 69.57 & 38.80 & 76.39 & 58.22 \\
Mmaba2-2.7B-PLU SiLU \& SoftPlus & 4.09 & 36.26 & 66.60 & 61.80 & 69.61 & 38.80 & 76.39 & 58.24 \\
\bottomrule
\end{tabular}
\end{table}

We evaluate XAMBA's effectiveness in reducing inference latency for Mamba-based models.
Figure~\ref{plot:result}(a) shows the average latency of a single-block Mamba-2 130M model on the Intel\textregistered\ Core\texttrademark\ Ultra Series 2 NPU. The CumBA technique reduces latency by $2.7 \times$ over the baseline, while ReduBA provides a $1.2 \times$ reduction. \textit{Reduced inference latency directly translates to improved Tokens/second.} Combining CumBA and ReduBA yields a $4.8 \times$ reduction, demonstrating the performance gains of XAMBA.
Figure~\ref{plot:result}(b) presents the normalized inference latency breakdown for the Mamba-2 130M model, comparing the baseline with the optimized CumBA technique. In the baseline, CumSum operations contribute over 50\% of the total latency. CumBA reduces this by transforming CumSum into matrix multiplication using a precomputed mask, achieving a $2.7 \times$ reduction in latency.
Figure~\ref{plot:result}(c) shows the first inference latency breakdown for the Mamba 130M model with ActiBA optimizations. Using PLU for SoftPlus in a piecewise linear fashion reduces latency by $1.2 \times$. Further reductions ($1.8 \times$) are achieved with SiLU, leading to a total $2.6 \times$ latency reduction with negligible quality loss when both SoftPlus and SiLU are mapped to PLU.
Table~\ref{tab:mamba_results} demonstrates that ActiBA's hardware-friendly approximations introduce only negligible quality tradeoffs. For the original Mamba architecture, the largest accuracy drop occurs with the 130M model (42.03\% $\to$ 40.67\%, a 1.36\% reduction), while larger models like the 1.4B version show remarkably stable performance (54.24\% $\to$ 54.15\%, just 0.09\% difference). The Mamba2 variants exhibit even greater robustness to approximations---the 130M model's accuracy remains virtually unchanged (42.52\% $\to$ 42.53\%), and interestingly, the 2.7B model shows a slight improvement (58.22\% $\to$ 58.24\%). These results collectively confirm that ActiBA's optimizations successfully maintain model quality, with the maximum observed degradation being well under 1.5\% even for the smallest model, while most variants experience changes of less than 0.1\%.
We set a KPI target of 50 Tokens/s, following MobileLLM-125M~\cite{mobilellm} (a comparable LLM), to ensure sufficient client-side responsiveness. With ActiBA optimizations, the Mamba-130M model's decoding performance increases from 100 Tokens/s to 260 Tokens/s, surpassing the 50 Tokens/s KPI target.
XAMBA’s optimizations, demonstrated on the 130M models, extend to larger models and inputs having similar bottlenecks, enabling comparable or greater speed-ups. While performance gains vary by model size and workload, the core principles hold. \textit{Ongoing work will further explore scalability and optimization for larger models.}

\section{Conclusion \& Future Work}\label{conclusion}
This work presents XAMBA, the first framework optimizing SSMs on COTS NPUs, removing the need for specialized accelerators. XAMBA mitigates key bottlenecks in SSMs like CumSum, ReduceSum, and activations using ActiBA, CumBA, and ReduBA, transforming sequential operations into parallel computations. These optimizations improve latency, throughput (Tokens/s), and memory efficiency. Future work will extend XAMBA to other models, explore compression, and develop dynamic optimizations for broader hardware platforms.

\subsubsection*{Acknowledgments}
This work was supported in part by the Center for the Co-Design of Cognitive Systems (CoCoSYS) and the Center on Cognitive Multispectral Sensors (CogniSense), two research centers under the Joint University Microelectronics Program (JUMP) 2.0, a Semiconductor Research Corporation (SRC) initiative sponsored by DARPA. The authors would like to thank Souvik Kundu (Intel Labs) and Zhifan Ye (Georgia Tech) for insightful discussions that contributed to this work.

\bibliography{xamba}

\begin{thebibliography}{18}
\providecommand{\natexlab}[1]{#1}
\providecommand{\url}[1]{\texttt{#1}}
\expandafter\ifx\csname urlstyle\endcsname\relax
  \providecommand{\doi}[1]{doi: #1}\else
  \providecommand{\doi}{doi: \begingroup \urlstyle{rm}\Url}\fi

\bibitem[Azizi et~al.(2025)Azizi, Kundu, Sadeghi, and Pedram]{mambaextend}
Seyedarmin Azizi, Souvik Kundu, Mohammad~Erfan Sadeghi, and Massoud Pedram.
\newblock Mambaextend: A training-free approach to improve long context extension of mamba.
\newblock In \emph{ICLR}, 2025.

\bibitem[Dao \& Gu(2024)Dao and Gu]{mamba2}
Tri Dao and Albert Gu.
\newblock {Transformers are SSMs: generalized models and efficient algorithms through structured state space duality}.
\newblock In \emph{ICML}, 2024.

\bibitem[Fei \& Abdelfattah(2024)Fei and Abdelfattah]{llm_npu}
Anthony Fei and Mohamed~S. Abdelfattah.
\newblock {NITRO: LLM Inference on Intel Laptop NPUs}, 2024.
\newblock URL \url{https://arxiv.org/abs/2412.11053}.

\bibitem[Gu \& Dao(2024)Gu and Dao]{mamba}
Albert Gu and Tri Dao.
\newblock {Mamba: Linear-Time Sequence Modeling with Selective State Spaces}.
\newblock In \emph{COLM}, 2024.

\bibitem[Gu et~al.(2020)Gu, Dao, Ermon, Rudra, and R{\'e}]{hippo}
Albert Gu, Tri Dao, Stefano Ermon, Atri Rudra, and Christopher R{\'e}.
\newblock {HiPPO: Recurrent Memory with Optimal Polynomial Projections}.
\newblock In \emph{NeurIPS}, 2020.

\bibitem[Gu et~al.(2022)Gu, Goel, and Re]{s4}
Albert Gu, Karan Goel, and Christopher Re.
\newblock {Efficiently Modeling Long Sequences with Structured State Spaces}.
\newblock In \emph{ICLR}, 2022.

\bibitem[Intel()]{openvino}
Intel.
\newblock {Intel\textregistered\ Distribution of OpenVINO\texttrademark\ Toolkit}.
\newblock URL \url{https://www.intel.com/content/www/us/en/developer/tools/openvino-toolkit/overview.html}.
\newblock Accessed: Feb. 07, 2025.

\bibitem[{Intel Corporation}(2024{\natexlab{a}})]{lnl}
{Intel Corporation}.
\newblock {Intel® Core™ Ultra series mobile processors product brief}, 2024{\natexlab{a}}.
\newblock URL \url{https://www.intel.com/content/www/us/en/products/docs/processors/core-ultra/core-ultra-series-2-mobile-product-brief.html}.
\newblock Accessed: Feb. 07, 2025.

\bibitem[{Intel Corporation}(2024{\natexlab{b}})]{mtl}
{Intel Corporation}.
\newblock {Intel® Core™ Ultra series 1 product brief}, 2024{\natexlab{b}}.
\newblock URL \url{https://www.intel.com/content/www/us/en/products/docs/processors/core-ultra/core-ultra-series-1-product-brief.html}.
\newblock Accessed: Feb 07, 2025.

\bibitem[Li et~al.(2024)Li, Huang, Xu, Liu, Ding, Xu, and Dai]{marca}
J.~Li, S.~Huang, J.~Xu, J.~Liu, L.~Ding, N.~Xu, and G.~Dai.
\newblock {MARCA: Mamba Accelerator with ReConfigurable Architecture}.
\newblock In \emph{arXiv}, 2024.
\newblock URL \url{https://arxiv.org/abs/2409.11440}.

\bibitem[Liu et~al.(2024)Liu, Zhao, Iandola, Lai, Tian, Fedorov, Xiong, Chang, Shi, Krishnamoorthi, Lai, and Chandra]{mobilellm}
Zechun Liu, Changsheng Zhao, Forrest Iandola, Chen Lai, Yuandong Tian, Igor Fedorov, Yunyang Xiong, Ernie Chang, Yangyang Shi, Raghuraman Krishnamoorthi, Liangzhen Lai, and Vikas Chandra.
\newblock {MobileLLM: optimizing sub-billion parameter language models for on-device use cases}.
\newblock In \emph{ICML}, 2024.

\bibitem[{OpenVINO Documentation}(2024)]{openvino_ops}
{OpenVINO Documentation}.
\newblock {OpenVINO {IR} Format: Operation Sets and Specifications}.
\newblock \url{https://docs.openvino.ai/2024/documentation/openvino-ir-format/operation-sets/operation-specs.html}, 2024.
\newblock Accessed: Feb. 07, 2025.

\bibitem[Patro \& Agneeswaran(2024)Patro and Agneeswaran]{mamba360}
Badri~Narayana Patro and Vijay~Srinivas Agneeswaran.
\newblock {Mamba-360: Survey of State Space Models as Transformer Alternative for Long Sequence Modelling: Methods, Applications, and Challenges}.
\newblock In \emph{arXiv}, 2024.
\newblock URL \url{https://arxiv.org/abs/2404.16112}.

\bibitem[Raha et~al.(2024)Raha, Mathaikutty, Ghosh, and Kundu]{flexnn}
Arnab Raha, Deepak~A. Mathaikutty, Soumendu~K. Ghosh, and Shamik Kundu.
\newblock {FlexNN: A Dataflow-aware Flexible Deep Learning Accelerator for Energy-Efficient Edge Devices}.
\newblock In \emph{arXiv}, 2024.
\newblock URL \url{https://arxiv.org/abs/2403.09026}.

\bibitem[Reggiani et~al.(2023)Reggiani, Andri, and Cavigelli]{flex_sfu}
Enrico Reggiani, Renzo Andri, and Lukas Cavigelli.
\newblock {Flex-SFU: Accelerating DNN Activation Functions by Non-Uniform Piecewise Approximation}.
\newblock In \emph{DAC}, 2023.

\bibitem[Rhu et~al.(2018)Rhu, O'Connor, Chatterjee, Pool, Kwon, and Keckler]{zvc}
Minsoo Rhu, Mike O'Connor, Niladrish Chatterjee, Jeff Pool, Youngeun Kwon, and Stephen~W. Keckler.
\newblock {Compressing DMA Engine: Leveraging Activation Sparsity for Training Deep Neural Networks}.
\newblock In \emph{HPCA}, 2018.

\bibitem[Wang et~al.(2018)]{rnn_npu}
X.~Wang et~al.
\newblock {Efficient Inference of Recurrent Neural Networks on Neural Processors}.
\newblock \emph{IEEE Transactions on Neural Networks and Learning Systems}, 2018.

\bibitem[Yang et~al.(2019)Yang, Wei, Tu, Zeng, Kinsy, Zheng, and Ren]{dse_act}
Tao Yang, Yadong Wei, Zhijun Tu, Haolun Zeng, Michel~A. Kinsy, Nanning Zheng, and Pengju Ren.
\newblock {Design Space Exploration of Neural Network Activation Function Circuits}.
\newblock \emph{IEEE Transactions on Computer-Aided Design of Integrated Circuits and Systems}, 2019.

\end{thebibliography}
\bibliographystyle{iclr2025_conference}

\appendix
\newpage
\section{Appendix}

\begin{figure}[b!]
\begin{center}
\includegraphics[width=0.65\columnwidth]{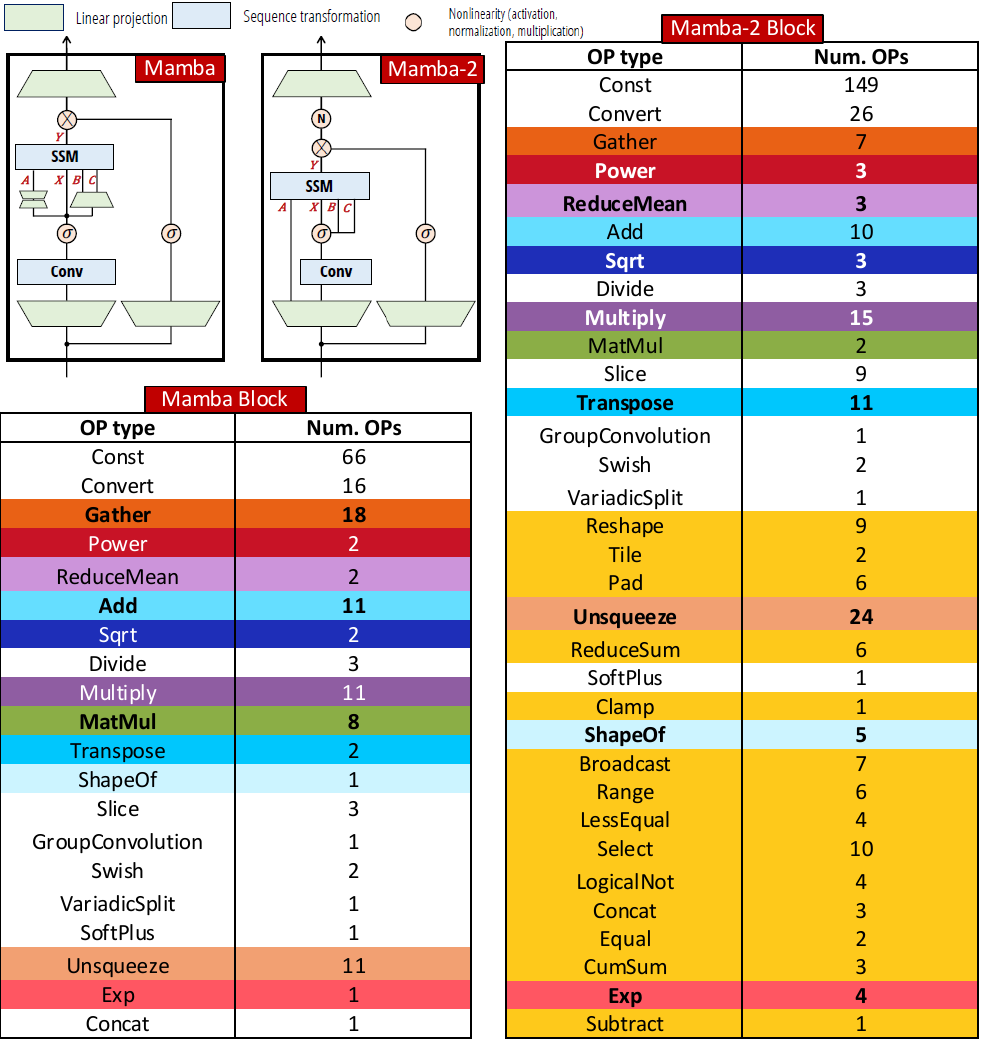}
\end{center}
\caption{Mamba~\cite{mamba} vs. Mamba-2~\cite{mamba2} showcasing structural differences and operational trade-offs. Mamba-2 simplifies the design but suffers from decreased performance on Intel\textregistered\ AI PCs~\cite{lnl}~\cite{mtl} due to less hardware-efficient operations.}\label{fig:mamba_vs_mamba2}
\end{figure}

\subsection{Mamba vs. Mamba-2}\label{subsec:mamba1_vs_2}
This section presents a comparative analysis of the Mamba~\cite{mamba} and Mamba-2~\cite{mamba2} architectures, focusing on their structural and operational distinctions. Fig.~\ref{fig:mamba_vs_mamba2} compares the architectures and operations of the Mamba and Mamba-2 blocks, highlighting their structural and computational differences. Mamba-1 focuses on leveraging structured state-space models (SSMs) for long-range sequence modeling through memory initialization using a high-order polynomial projection operator (HiPPO)~\cite{hippo}, a selection mechanism, and hardware-aware computing. It employs a sequential processing structure where linear projections are applied to the input sequence and state-space parameters $[A, B, C]$ in stages, with selective scanning through the SSM unit to map inputs $X$ to outputs $Y$. The Mamba-1 block relies on skip connections to reuse features and mitigate performance degradation during training. Mamba-2 simplifies this design by introducing the structured state-space duality (SSD) framework, which connects SSM and attention mechanisms. Instead of sequential linear projections, the SSD layer processes $[X, A, B, C]$ simultaneously using a single projection, akin to generating $Q, K, V$ in standard attention mechanisms. This reduces computational overhead and accelerates processing. Additionally, Mamba-2 adds a normalization layer after the skip connection to improve training stability. 

Fig.~\ref{fig:mamba_vs_mamba2} also highlights the operator~\cite{openvino_ops} differences between Mamba and Mamba-2, emphasizing the trade-offs in performance on Intel\textregistered\ AI PCs~\cite{lnl}~\cite{mtl}. Mamba-2 introduces new operators like \texttt{CumSum} and \texttt{ReduceSum}, while reducing gather operations (from 18 to 7). However, computationally expensive operations such as \texttt{power} and \texttt{sqrt} increase, and data-parallel operations critical for DPUs, like \texttt{MatMul} (reduced from 8 to 2) and \texttt{Add} (from 11 to 10), are decreased. These changes negatively impact Mamba-2's performance on Intel\textregistered\ AI PCs, as it shifts away from hardware-optimized operations. Despite structural simplifications through its SSD framework, Mamba-2 performs worse than Mamba on AI PCs due to increased reliance on less hardware-efficient operators.

\end{document}